\documentclass[twoside,twocolumn]{article}

\usepackage{fitee}
\usepackage{gensymb}

\usepackage[ hidelinks, breaklinks = true]{hyperref}  
\addto{\captionsenglish}{%
  
}

\begin{document}

\title{A selected review on reinforcement learning based control for autonomous underwater vehicles}

\author{Yachu HSU}
\author{Hui WU}
\author{Keyou YOU$^{\ddagger}$}
\author{Shiji SONG}

\affil{\it\footnotesize Department of Automation and BNRist, Tsinghua University, Beijing 100084, China}

\shortauthor{Hsu et al.}	

\authmark{}



\corremailA{\{xuyz17, wuhui14\}@mails.tsinghua.edu.cn}
\corremailB{\{youky, shijis\}@tsinghua.edu.cn}


\abstract{Recently, reinforcement learning (RL) has been extensively studied and achieved promising results in a wide range of control tasks. Meanwhile, autonomous underwater vehicle (AUV) is an important tool for executing complex and challenging underwater tasks. The advances in RL offers ample opportunities  for developing intelligent AUVs. This paper provides a selected review on RL based control for AUVs with the focus on applications of RL to low-level control tasks for underwater regulation and tracking. To this end, we first present a concise introduction to the RL based control framework. Then, we provide an overview of RL methods for AUVs control problems, where the main challenges and recent progresses are discussed. Finally, two representative cases of RL-based controllers are given in detail for the model-free RL methods on AUVs. }


\clc{TP}

\inpress	

\publishyear{2019}
\vol{19}
\issue{1}
\pagestart{1}
\pageend{5}

\support{This work was supported in part by the National Key Research and Development Program of China under Grant No.2016YFC0300801, and National Natural Science Foundation of China under Grants No.41576101 and No.41427806.}

\orcid{Ke-you YOU, https://orcid.org/0000-0003-4355-5340}	
\articleType{Review:}

\maketitle

\section{Introduction} \label{sec:introduction}
The development of AUVs was initially motivated by the desire to explore the Arctic waters in 1957. Since then, AUVs have received considerable attention and intensive efforts have been made to deploy AUVs in various underwater environments. These versatile vehicles bring a revolution to the field of ocean research. The development of controllers also contributes to this as it is also key to the capability of AUVs. Many controllers have been designed for AUVs to complete manifold military and civilian tasks, including source seeking \citep{li2018auv}, pipeline inspecting \citep{xiang2010coordinated}, seafloor mapping \citep{ribas2011girona}, etc.

Among them, RL-based controllers are highly anticipated for their capability to enable adaptive autonomy in an optimal manner \citep{kiumarsi2017optimal}. RL algorithms provide control policies that maximize the quantitative performance throughout a well-designed task by learning from ongoing interactions with the environment. RL looks ahead to future events and focuses on long-term performance, making it appealing to control problems. In the control fields of unmanned aerial vehicles (UAVs) and unmanned ground vehicles (UGVs), RL algorithms are widely studied. The successes of \citet{waslander2005multi}, \citet{kim2004autonomous} and \citet{bagnell2001autonomous} demonstrate that RL-based controllers perform better than classic controllers or highly trained pilots. \citet{abbeel2010autonomous} presented apprenticeship learning algorithms that allowed autonomous helicopter to perform arbitrary challenging aerobatic maneuvers. \citet{hester2011real} conducted real-time learning on a physical vehicle to control its velocity through pedals. The velocity was accurately tracked after 3 minutes. \citet{kendall2019learning} realized autonomous driving via RL where the full sized vehicle learnt to follow lanes from scratch within 30 minutes using on-board computers.

The success of RL in ground and aerial vehicle control community suggests its potential for controlling AUVs. In fact, RL framework has been introduced to achieve persistent autonomy and precise control for AUVs. This paper briefly surveys the progress of the implementation of RL on different low-level control tasks for AUVs, hoping to inspire more research into the application of RL.

The remainder of this paper is organized as follows. Section 2 describes basic concepts in RL with an introduction of RL algorithms. Section 3 discusses challenges in applying RL to control AUVs. Recent studies in this area are briefly introduced in Section 4. In Section 5, the cases of two RL-based controllers for different underwater tasks are presented in detail to better illustrate the advantages of RL-based controllers. Finally, a conclusion is made in Section 6.

\section{Basics of RL}
This section concisely introduces basic concepts and foundational algorithms of RL. Due to the uncertainty of underwater dynamics, this section mainly focuses on model-free algorithms.

\subsection{Markov decision process}
This subsection is mainly based on \citet{sutton2018reinforcement}. Formally, RL aims to solve the Markov decision process(MDP) based problem, which consists of four basic elements: a set of valid states $\mathcal{S}$ (and the starting state distribution $\rho_{0}$), a set of valid actions $\mathcal{A}$, reward function $r(s,a)$ and transition probability function $p\left(s^{\prime}, r | s, a\right)$. Transitions should depend only on the most recent state and action, which is known as the Markov property.

\begin{figure}[tbh]
	\centering
	\includegraphics[scale=0.15]{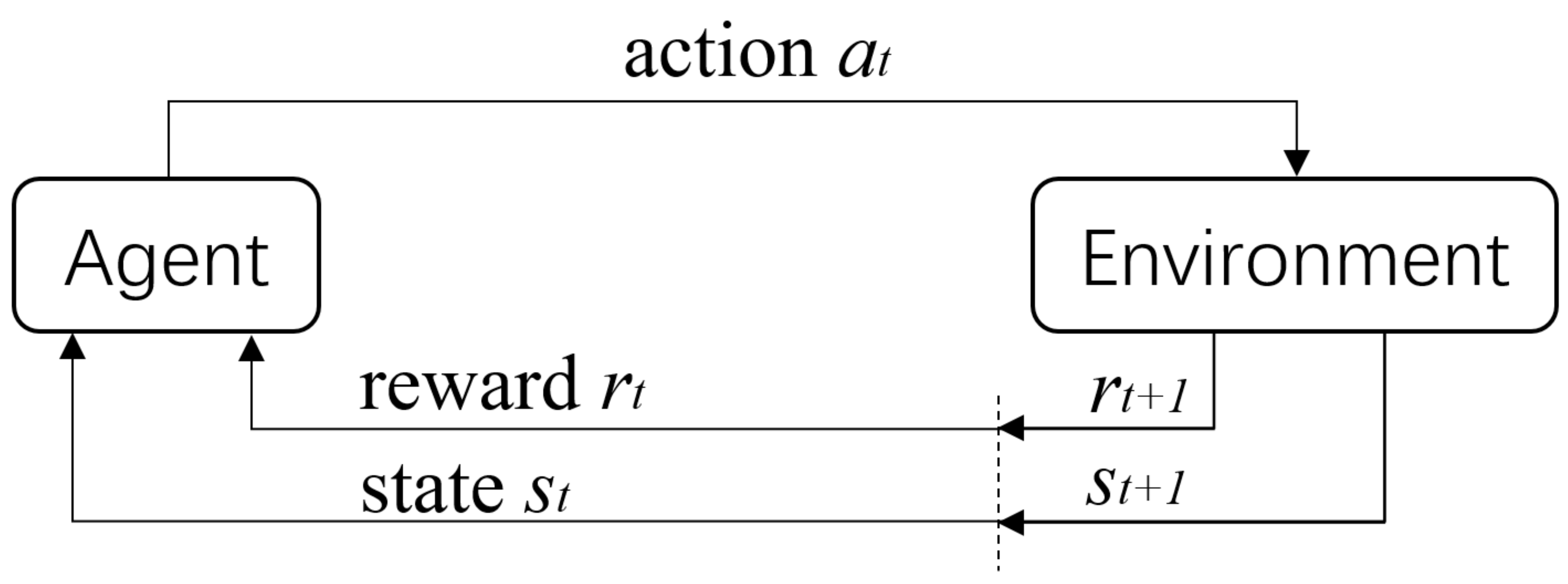}
	\caption{The agent-environment interaction.}
\label{fig:mdp}
\end{figure}

By adopting RL, an agent learns the mapping between situations and control outputs from interactions with the environment, so as to optimize its control performance. Learning from these repeated interactions enables RL to handle the case where dynamic programming \citep{bertsekas1995dynamic} is not applicable, i.e., the case when the function $p$ is unknown. At every step of the interaction, the agent observes a state and chooses an action based on the observation. The environment then transits to a new state depending on the current state and the action just made. A reward signal is received to evaluate such an action. Fig.~\ref{fig:mdp} illustrates this agent-environment interaction, which gives rise to a sequence of states, actions and rewards $\tau=\left(s_{0}, a_{0}, r_{1}, s_{1}, a_{1}, r_{2}, \dots\right)$. The agent's goal is to maximize the return $R(\tau)=\sum_{t=0}^{\infty} \gamma^{t} r_{t}$, namely the cumulative reward received during the interaction period, where $\gamma \in(0,1]$ is a discount factor to assign decayed weights to the future rewards.

\subsection{Value-based RL}
The rule which an agent used for choosing actions is called a policy. The expected return received when starting in a state $s$ under a specific policy $\pi$ is called the value at $s$. For MDPs, we define the state-value function and state-action-value function as
\begin{align}\label{value}
  V^{\pi}(s) & ={\mathrm{E}}_{\tau \sim \pi}\left[R(\tau) | s_{0}=s\right] \\
  Q^{\pi}(s, a) & ={\mathrm{E}}_{\tau \sim \pi}\left[R(\tau) | s_{0}=s, a_{0}=a\right]
\end{align}
where $\tau$ denotes a trajectory and $s_{0}, a_{0}$ are its starting state and action. The state-action-value $Q^{\pi}(s, a)$ can construct an optimal policy by choosing an action that maximizes the value function. A policy is optimal if it can reach the highest expected return for all states, while a value function is optimal if it acts according to the optimal policy. The optimal value functions can be defined as
\begin{align}\label{optimal value}
  V^{*}(s) & =\max_{\pi} {\mathrm{E}}_{\tau \sim \pi}\left[R(\tau) | s_{0}=s\right], \\
  Q^{*}(s, a) & =\max_{\pi} {\mathrm{E}}_{\tau \sim \pi}\left[R(\tau) | s_{0}=s, a_{0}=a\right].
\end{align}

Value functions play an important role in designing RL algorithms, because they evaluate how well a state or state-action pair is and guide the algorithm to search an optimal policy. Methods used to estimate value functions are typically derived from Bellman equations
\begin{align}\label{Bellman}
  V^{\pi}(s) & ={\mathrm{E}}_{s^{\prime} \sim P}\left[r(s, a)+\gamma V^{\pi}\left(s^{\prime}\right)\right] \\
  Q^{\pi}(s, a) & ={\mathrm{E}}_{s^{\prime} \sim P}\left[r(s, a)+\gamma {\mathrm{E}}_{a^{\prime} \sim \pi}\left[Q^{\pi}\left(s^{\prime}, a^{\prime}\right)\right]\right. \\
  V^{*}(s) & =\max _{a} {\mathrm{E}}_{s^{\prime} \sim P}\left[r(s, a)+\gamma V^{*}\left(s^{\prime}\right)\right] \\
  Q^{*}(s, a) & ={\mathrm{E}}_{s^{\prime} \sim P}\left[r(s, a)+\gamma \max _{a^{\prime}} Q^{*}\left(s^{\prime}, a^{\prime}\right)\right]
\end{align}
where $P$ denotes state transition probabilities $p\left(s^{\prime}, r | s, a\right)$. Most model-free RL procedures can be abstracted as alternating between policy evaluation and policy improvement. The former estimates the value function of the current policy and the other improves the policy with respect to the estimated value function, usually by making it greedy to the current value function. This process stabilizes when the policy is greedy to its own value function, which matches the Bellman equations of the optimal value and policy. This idea is referred to as generalized policy iteration (GPI).

The self-consistency Bellman equations imply that the estimation of value functions can be improved by bootstrapping. Inspired by this idea, temporal-difference (TD) learning is a kind of commonly used value-based RL algorithm. It updates the estimation by minimizing the TD error $\delta$, i.e.,
\begin{align}\label{TD}
  Q^{\pi}\left(s_{t}, a_{t}\right) & \leftarrow Q^{\pi}\left(s_{t}, a_{t}\right)+\alpha \delta
\end{align}
where $\alpha$ denotes the learning rate and $\delta=Y-Q^{\pi}\left(s_{t}, a_{t}\right)$ is the error between the target value Y and its estimated value. The target varies in different algorithms, for example $Y=r_{t}+\gamma Q^{\pi}\left(s_{t+1}, a_{t+1}\right)$ in an on-policy algorithm SARSA and $Y=r_{t}+\gamma \max _{s} Q^{\pi}\left(s_{t+1}, a\right)$ in an off-policy algorithm Q-learning. On-policy means the updated policy is consistent with the policy used for sampling, whereas off-policy algorithm uses a different policy for interacting with the environment.

Beside bootstrapping, value functions can also be updated by Monte Carlo (MC) methods in episodic tasks. In these methods, value functions are estimated by averaging the returns observed so far. MC methods and TD methods are like two extremes and can be unified by taking n-step bootstrapping. This usually performs better as if it utilized the advantages of previous methods, whereas more computation is also required.

\subsection{Policy-based RL}
The major defect of value-based algorithms is that the maximum operation makes them inapplicable for the continuous action space. A solution is to use parameterized policies. Unlike greedy policies, a parameterized policy $\pi_{\theta}$ can be stochastic and thereby is more suitable for problems with imperfect information. Moreover, by utilizing the neural network as a non-linear function approximator along with some modifications to stabilize learning, the policy is able to handle high-dimensional observations. The performance of the policy $J\left(\pi_{\theta}\right)$ can be defined as either the expectation of cumulative discounted reward or average reward, for example
\begin{align}\label{performance}
  J\left(\pi_{\theta}\right)= V^{\pi_{\theta}}(s_0).
\end{align}
Then the parameter can be directly optimized for the performance by gradient descent, i.e.,
\begin{align}\label{gradient descent}
  \theta_{k+1}=\theta_{k}+\alpha \nabla_{\theta} J\left(\pi_{\theta_{k}}\right).
\end{align}
Policy Gradient Theorem in \citet{sutton2000policy} lays the foundation for these algorithms
\begin{align}\label{policy gradient theorem}
  \nabla_{\theta} J\left(\pi_{\theta}\right)= E_{s \sim \rho^{\pi}, a \sim \pi_{\theta}}\left[\nabla_{\theta} \log \pi_{\theta}(a | s) Q^{\pi}(s, a)\right]
\end{align}
where $\rho^{\pi}$ denotes the state distribution. It shows that the gradient of the performance function with respect to the policy parameter can be expressed by an expectation without concerning the effect of policy changes on the state distribution. This gradient can be estimated by sampling methods as long as an unbiased expectation is guaranteed. REINFORCE is a kind of straightforward Monte Carlo implementation, which estimates the action-value function with sampled returns \citep{williams1992simple}.

Another family of variants is the famous actor-critic methods. They take the value function estimation mentioned above as the critic part of the algorithm to better learn the policy parameter. The policy learning part referred to as the actor consults the estimated value function $Q^{\omega}(s, a)$ when computing the performance gradient. It is proved that under certain constraints, this approximation will not results in bias \citep{sutton2000policy}
\begin{align}\label{policy gradient theorem with approximation}
  \nabla_{\theta} J\left(\pi_{\theta}\right)=\mathbb{E}_{s \sim \rho^{\pi}, a \sim \pi_{\theta}}\left[\nabla_{\theta} \log \pi_{\theta}(a | s) Q^{\omega}(s, a)\right].
\end{align}

It is obvious that when the variance of the stochastic policy is zero, the policy reduces to a deterministic policy. \cite{silver2014deterministic} extended the policy gradient framework to deterministic policies by proving that as the variance approaches zero, the stochastic policy gradient converges to deterministic gradient with the following form
\begin{align}\label{deterministic policy gradient theorem}
  \nabla_{\theta} J\left(\mu_{\theta}\right)=\mathbb{E}_{s \sim \rho^{\mu}}\left[\nabla_{\theta} \mu_{\theta}(s) \nabla_{a} Q^{\mu}\left.(s, a)\right|_{a=\mu_{\theta}(s)}\right].
\end{align}
This adaption can also be interpreted as approximating the maximization with the policy
\begin{align}
    \max _{a} Q(s, a) \approx Q(s, \mu(s)).
\end{align}
Deterministic policy gradient (DPG) needs not to integrate over the action space, which is appealing when the dimension of the action space is high. Popular algorithms derived from this idea have already proved their advantages \citep{lillicrap2015continuous, fujimoto2018addressing}.

Normal policy gradient methods measure the distance between policies in parameter spaces, yet it is better to measure the distance on the probability manifold to ensure the performance improvement in every step. This is the basic idea of the natural policy gradient \citep{amari1998natural}. These methods optimize policy with different surrogate objective functions \citep{schulman2015trust, schulman2017proximal}.

\subsection{Model-based RL}
Model-free RL learns directly from interactions, whereas model-based RL rely on a model of the environment, which can be described as the transition probability function $p\left(s^{\prime}, r | s, a\right)$. The model accuracy has a great impact on the performance of the model-based RL. The policy may perform poorly in the real test when the learnt model is inaccurate. Two main concerns in model-based RL are the way to obtain models and how to utilize them.

In some simple tasks, an accurate mathematical model can be established based on the priori knowledge. For other cases, it is possible to learn a model from interactions with the environment. Provided that assumptions about the model are given, the unknown parameters of the model can be inferred by methods like linear regression. If there is no prior knowledge of the model, a common approach for learning the model is the Gaussian Processes (GP), which can build a distribution over the transition function \citep{polydoros2017survey}.

Once the model of the environment is accessible, then a possible step or even all possible episodes can be generated for planning methods like dynamic programming, heuristic search and exhaustive search. There are also many other ways of combining models with model-free algorithms, e.g., regarding the planning method as an expert that the policy should learn from \citep{anthony2017thinking}, considering plans as side information for the policy \citep{racaniere2017imagination}, producing simulated experiences for data augmentation \citep{feinberg2018model}, etc.

Recently, a rich class of RL methods has been extensively studied, although a further review on the state of the art in RL will not be included in this paper. Instead, we will focus on integrating RL into the control of AUVs in the following sections.

\section{Challenges for implementing RL to control AUVs}
\begin{figure}[tbh]
	\centering
	\includegraphics[scale=0.15]{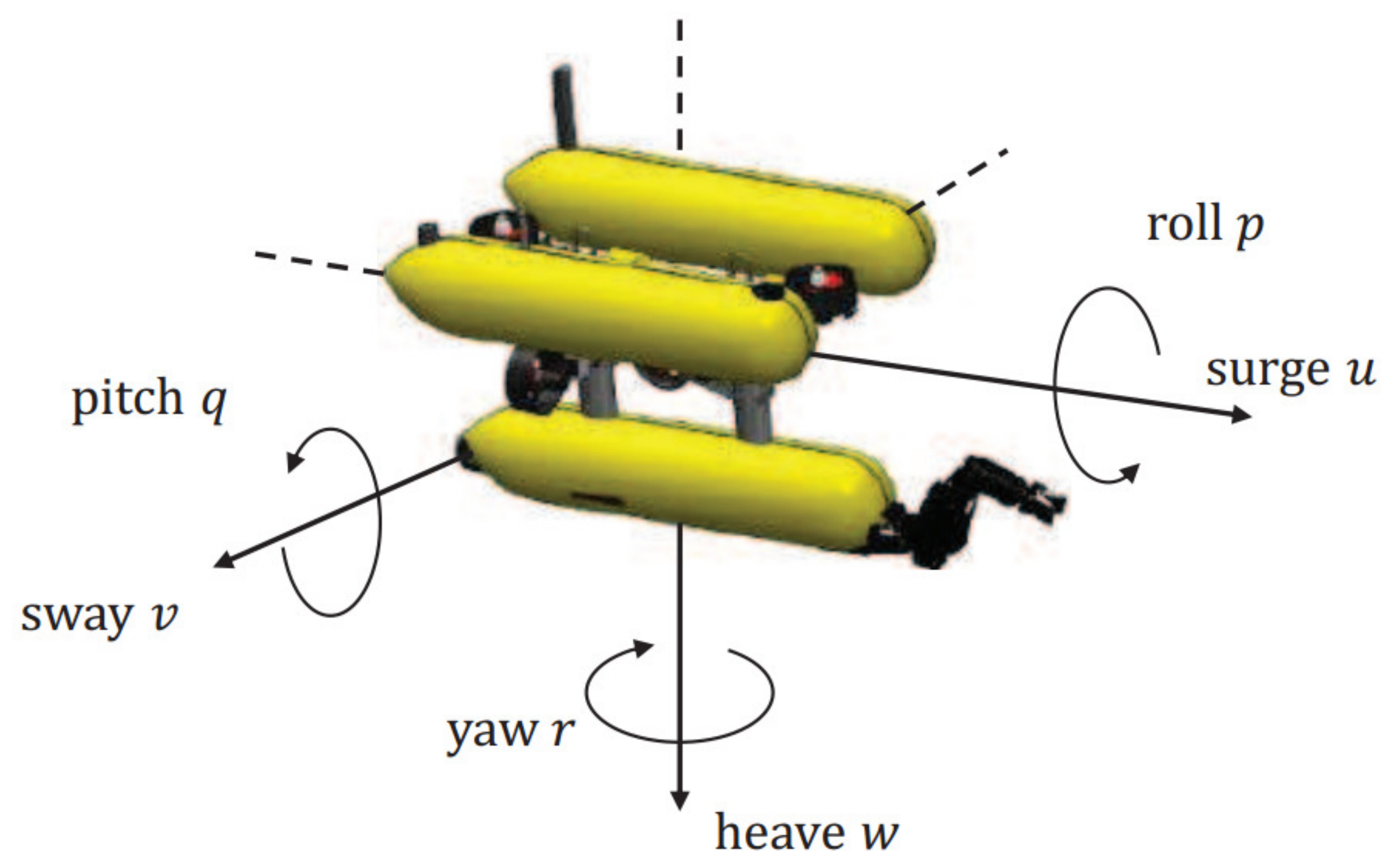}
	\caption{The six DOFs(degrees of freedom) motions of the AUV.}
\label{fig:dof}
\end{figure}

The fact that RL-based control methods can learn form interactions distinguishes them from classic control methods including PID control \citep{khodayari2015modeling}, adaptive control \citep{narasimhan2006adaptive}, backstepping control \citep{lapierre2007nonlinear}, sliding-mode control \citep{elmokadem2016trajectory}, etc. Controlling an AUV is kinematically similar to the problem of controlling a free-floating rigid body in a six-dimensional space (Fig.~\ref{fig:dof}). However, the underwater environment complicates the dynamics \citep{antonelli2018underwater}. For most classic controllers, one of the major obstacles is the lack of an accurate dynamic model for designing controllers. Moreover, the models are manually decoupled or linearized without a sufficient consideration on the uncertainties and disturbances, which are important environmental underwater characteristics. On the other hand, RL-based controllers can be trained without a dynamic model while enabling adaptive autonomy. Challenges of controlling AUVs in the context of RL differ in many ways. Some possible challenges are discussed in this section.

\subsection{Sample efficiency}
To apply RL algorithms to AUVs, experiences need to be acquired by interacting with the physical system. It is obvious that carrying out such an experiment with an AUV is costly in terms of time, labour and finances. More specifically, an AUV is expensive to build and needs careful maintenance to reduce wear as well as avoid crashing. Meanwhile, whether to build a water tank or to find a site with suitable underwater environment is not an easy task. Even if all preparation works are ready, the process to collect data itself is time consuming. Therefore gaining a better sample efficiency for minimizing interactions has become an outstanding issue, outweighing limiting memory consumption and computational complexity. Off-policy methods are more suitable in this case since they are able to reuse the experiences collected, namely more sample efficient. Model-based methods are widely used in robot control for their promise of sample efficiency, though models of AUVs and underwater environment are often poorly known.

\subsection{Tradeoff between exploration and exploitation}
During the learning process, exploiting means to select greedy actions, which have the greatest estimated value. This maximizes the expected reward on the one step, whereas exploring by selecting nongreedy actions may produce the greater total reward in the long run \citep{sutton2018reinforcement}. Additionally, by taking these actions, more kinds of states will be visited, which results in a more robust policy.

Guaranteeing sufficient exploration has been a long existing problem in RL algorithms, and is especially important in the control of AUVs to provide robustness to the variable underwater environment. Although learning from mistake is beneficial in most cases, exploring underwater with an AUV is rather complicated. The price of damaging an AUV is particularly high, considering the cost, physical labour and long waiting period for repairing an AUV. Safe exploration is a key issue for practical application of RL algorithms, which is often neglected in the general RL community \citep{kober2013reinforcement}. One possible solution is to prepare emergency protocol which has a higher priority when AUVs encounter danger, such as getting too closed to an obstacle \citep{el2013two}.

\subsection{Model uncertainty}
The mathematical model of the dynamics of underwater vehicles is usually derived from Newton-Euler equations of a rigid body, in which the effect of inertial generalized forces, hydrodynamics, gravity, buoyancy and thrusters' presence are taken into account \citep{antonelli2018underwater}. However, most of these effects are either complex itself with high nonlinearity and time-varying characteristic or closely related to the exact structure of the AUV. In a word, it is hard to develop a reliable model for an AUV. Although learning with an accurate model can solve the problem in collecting real-world samples, bias in the model may cause sub-optimal or terrible performance in the real environment no matter how well the policy behaves when with the approximation model. The unexpected poor performance may cause irreversible damage to AUVs. For model-based algorithms and model-free algorithms that trained in a simulator, it is essential to deal with this reality gap issue. Compromise has to be made between accuracy and robustness. Under such a condition, the best policy should be the one that is robust to noises rather than the one with the highest reward.

\subsection{Partially observed state}
Most existing pure RL algorithms are designed under the assumption that the environment can be totally observed. Whereas, the underwater environment is noisy and uncertain, creating great challenges for AUVs to collect useful information.

For visual inspections in underwater domain, both optical and sonar systems are widely used \citep{ferreira2016underwater}. Optical sensors are expected to obtain high resolution data with helpful colour information. However, in turbid water, water molecules, dissolved organic and inorganic matter, and various types of suspended particles cause scattering and absorption of light, and results in dark and low contrast underwater images with poor visibility. Colour distortion also occurs because of the different attenuation rate inversely proportional to the wavelength of light \citep{lu2015contrast}. Sonars are able to look further, while providing low resolution images insufficient for object identification.

As for underwater localization, unlike UAVs and UGVs, radio or spread-spectrum communications and global positioning are disabled due to the rapid attenuation of higher frequency signals. Although acoustic-based sensors and communications can support the localization of AUVs, they are constrained by limited and distance-dependent bandwidth, time-varying multi-path propagation and low speed of sound \citep{heidemann2012underwater}. In many cases, control designers of AUVs need to balance the need to inspect the environment and the cost of higher energy consumption which diminished the available mission time. In a real-world task, only limited information can be gained, not to mention the quality of the information.

\section{RL applications in control of underwater vehicles}
This section briefly introduces recent works on controlling AUVs with RL. There are also works that focus on high-level decision tasks such as path planning \citep{yoo2016path, wang2018reinforcement, hu2019plume}, which do not involve the low level control of AUVs. An extension of this topic is beyond the scope of this paper.

\subsection{Modeling of MDPs}
Before applying RL methods, it is vital to frame the control task as a MDP. As mentioned above, there are four elements that have to be well defined. For the low level control problem, the actions are control inputs of AUVs and the transition probability is usually unaccessible. Hence the key is to find a proper state representation and design an appropriate reward function.

\subsubsection{State representations}
It is natural to define the raw observations as the state, since this is the most informative form. However, RL methods suffer from the curse of dimensionality. Larger amount of samples and computation are desired to ensure the convergence as the number of state-space dimension grows. Thus a proper formulation should involve fewer variables while avoiding perceptual aliasing, which is the case that different states cannot be distinguished by the given information. \citet{wu2018depth} considered three depth control problems precisely and designed the states carefully. More details about this piece of work will be introduced in the following section. Meanwhile, a good state representation can greatly improve the robustness of an algorithm, results in a more generalized policy, e.g., the goal oriented control architecture used in \citet{carlucho2018adaptive, carlucho2018auv} can omit the continuous retraining step when changing to a new goal.

\subsubsection{Design of reward functions}
On the other hand, the design of reward functions also holds great importance. When facing conflicting objectives, a simple treatment is to unify them with prescribed weights. This kind of scalarized reward functions are easy to implement and can produce a single optimal solution. Generally, a reward function for AUV control tasks consists of two parts, i.e., terms to evaluate the error and terms to restrict the thruster usage or penalize sudden changes. \citet{carlucho2018adaptive} illustrated the significance of each term. The former terms ensure that the AUV achieves the control target while the latter terms prevent the thruster outputs from violent oscillation. The weights are usually chosen empirically according to the relative importance of the objectives. \citet{yu2017deep} imposed constraints on weights, which were derived from Lyapunov theory to guarantee the stability of the control system. Nevertheless, the weights are still hard to tune, since even small changes in weights may cause a great difference in the learnt policy. \citet{ahmadzadeh2014multi} employed multi-objective RL that can discover multiple optimal solutions which satisfied different objectives respectively (shortest path, minimum final velocity and minimum heading error). An additional algorithm is then needed for selecting the optimal solution.

Piecewise function is also a common form of the reward function, in which different levels of reward will be given according to the preference for the current state. \citet{carlucho2018auv} suggested that by gradually tighten the definition of the preferred state, the agent should obtain more useful experiences. For example, if a positive reward signal can be obtained when reaching an spherical neighborhood of a desired way-point, then it is beneficial to decay the radius of the sphere throughout learning.

\subsection{RL for screw-driven underwater vehicles}
\begin{figure}[t!]
	\centering
	\includegraphics[scale=0.15]{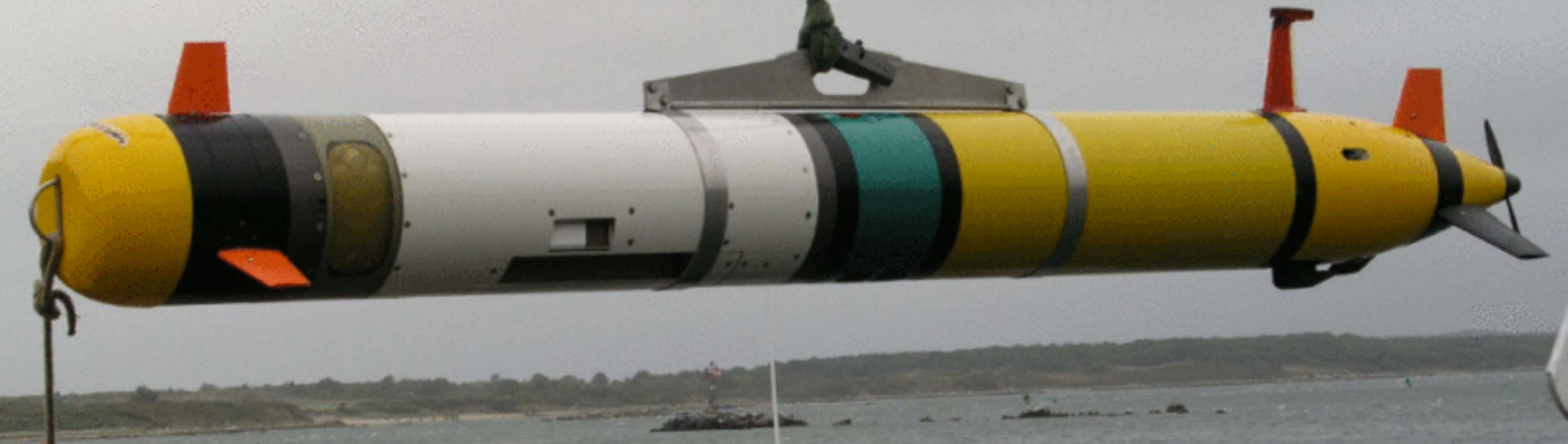}
	\caption{REMUS, a kind of screw-driven AUV \citep{stokey2005development}.}
\label{fig:remus}
\end{figure}

Classical AUVs are usually controlled by rotary propellers and control surfaces such as rudders and sterns. Most of them have the shape of a torpedo for hydrodynamic performance (Fig.~\ref{fig:remus}). Amounts of experiments have been carried out on them.

\subsubsection{Set-point regulation}
Stabilization is the most fundamental control task for AUVs. \citet{fernandez2014effect} conducted experiments on the speed control problem using Continuous Action-Critic Learning Automaton (CACLA). Unlike other policy gradient methods, CACLA only updated the policy in action space when the critique was strictly positive. Besides, it was proved that starting the training with outputs of PID as replacements of random actions helped to bias the learnt policy towards the optimal policy. \citet{walters2018online} carried out the regulation task in reality utilizing model-based dynamic programming. The dynamic model was learnt on-the-fly. They focused on the influence of the time-varying irrational current and presented the Lyapunov-based stability analysis to guarantee the convergence to the target state and optimal polices. \citet{carlucho2018adaptive} contributed to this field by conducting control tests of a real AUV on all six DOFs. The proposed deep RL algorithm was based on deep deterministic policy gradient(DDPG) and framed in the goal oriented control architecture.

\subsubsection{Way-point tracking}
Several attempts have been done for applying RL to the way point tracking problem for AUVs, which can be seen as a transition task between the station keeping task and tracking task. \citet{frost2014evaluation} presented a simplistic implementation of tabular Q-learning in both simulated and real scenario. The problem was discretized into a grid-world learning problem. Other studies all tried to discover fault-tolerant strategies, i.e., methods that can operate under thruster failure. This means that these algorithms should be able to control both over-actuated and under-actuated AUVs. \citet{carlucho2018auv} applied an algorithm similar to the one in \citet{carlucho2018adaptive}. \citet{jamali2014covariance} aimed at improving the robustness of the policy found by model-based direct policy search. A Gaussian noise was added to the inputs of the thrusters to test the sensitivity of the policy to noise. The results showed that the relationship between the performance in the noiseless setting and the robustness of the policy is unpredictable. Covariance analysis was used to measure the robustness in order to find a policy that performed well whilst being robust to noise.

Researchers of Istituto Italiano di Tecnologia carried out a series of experiments on this topic \citep{leonetti2013line, ahmadzadeh2014online, ahmadzadeh2014multi}. In 2013, an on-line controller framed within model-based policy search was proposed. Although the feasibility of the method was tested in simulator, the policy cannot be applied to real open water scenario for being an open-loop function of time. In 2014, this was solved by closing the loop with state feedbacks and the learnt policy was evaluated on a real AUV. Different levels of thruster failure was also considered. Nevertheless, the fact that the presented method requires the dynamic model of the AUV as well as related hydrodynamic parameters make it less appealing in practical use.

So far, only learnt policies have been tested, the test of learning a fault-tolerant policy online has not yet been performed in reality.

\subsubsection{Trajectory tracking}
Varying degrees of success have been achieved in the tracking control of AUVs. \citet{palomeras2012cola2} presented a control architecture for AUVs in which the RL algorithm was programmed in the reactive layer and tested in a real-time autonomous underwater task. A visual based cable tracking task was completed after applying a two-step learning process using natural actor-critic algorithm. The location and rotation of the cable were computed for two RL controllers to learn uncoupled policies for the yaw and sway action. The controllers were trained in the simulator before learning in reality to enhance the convergence rate. \citet{carlucho2018adaptive} adopted a similar learning strategy when conducting the velocity control task. In \citet{el2013two}, more real world experiments on cable tracking were conducted. To study the robustness of learnt policies, the policies were tested with different cable configurations without retraining. Another test changed the altitude of the AUV with respect to the cable during the online learning process. The results showed that the policies were with high adaptation capabilities.

Due to a variety of restrictions, unlike the station keeping task, other algorithms proposed for tracking are not yet sufficiently validated in real scene. \citet{sun2015target} used regularized extreme learning machine to replace the look-up table in Q-learning. However, the description of experiment settings and results was ambiguous. \citet{shi2018high} modified the calculation of the target value used for updating the critic in deterministic policy gradient algorithm. The so called pseudo averaged Q-learning method averaged over several previously learnt action-value estimations and benefited from multiple actors. This scheme stabilized the learning process by reducing the variance of target approximation error. In \citet{shi2018multi}, the proposed multi pseudo Q-learning based deterministic policy gradient algorithm employed multiple critics and multiple actors simultaneously. The critics were updated by the expected absolute Bellman error to accelerate the learning process.

To demonstrate the effectiveness of the proposed algorithms, some researchers gave out rigorous theoretical analysis on the stabilization of the control system. \citet{yu2017deep} solved the tracking problem through DDPG. The system was mathematically proved to be stable as long as the reward was chosen according to the Lyapunov stability principle. In 2014, \citet{cui2014neural} proposed a partially model-based adaptive control algorithm framed within the actor-critic architecture. The actor network compensated the uncertainties in dynamics and the critic network evaluated the tracking performance. In 2017, the input nonlinearities was considered in the dynamic model \citep{cui2017adaptive}. The nonlinearities included the actuator dead-zone and saturation as well as the relationship between the nominal and actual force. In 2019, the actor-critic adaptive control algorithm was further investigated for continuous-time systems with completely unknown dynamics \citep{guo2019integral}. A Nussbaum-type function was used to resolve unknown control directions. Compared with the previous algorithms in the discrete time manner \citep{cui2017adaptive}, it successfully avoided chattering of control inputs in steady-state phase. The other simulation showed its ability to gain results competitive to general neural network control which had access to the input dynamics. On the other hand, in \citet{guo2019event}, an event-triggered RL-based adaptive tracking control algorithm was investigated to reduce the update frequency of the controller. The algorithm was designed to consider the long-term performance index, unmodeled dynamics, and external disturbances simultaneously. Compared to the ordinary time-triggered methods, it significantly reduced the computational load and energy consumption.

\subsection{RL for bionic underwater vehicles}
Bionic AUVs, which mimic the swimming motions of underwater creatures, are developed to meet the higher requirements on endurance, system noise and especially maneuverability. Compared with the screw-driven AUVs, more efforts in control algorithms have to be made for these AUVs to acquire an optimal swimming pattern for the complicated dynamics.

\begin{figure}[!htb]
\centering
\begin{tabular}{cc}
\includegraphics[scale=0.1]{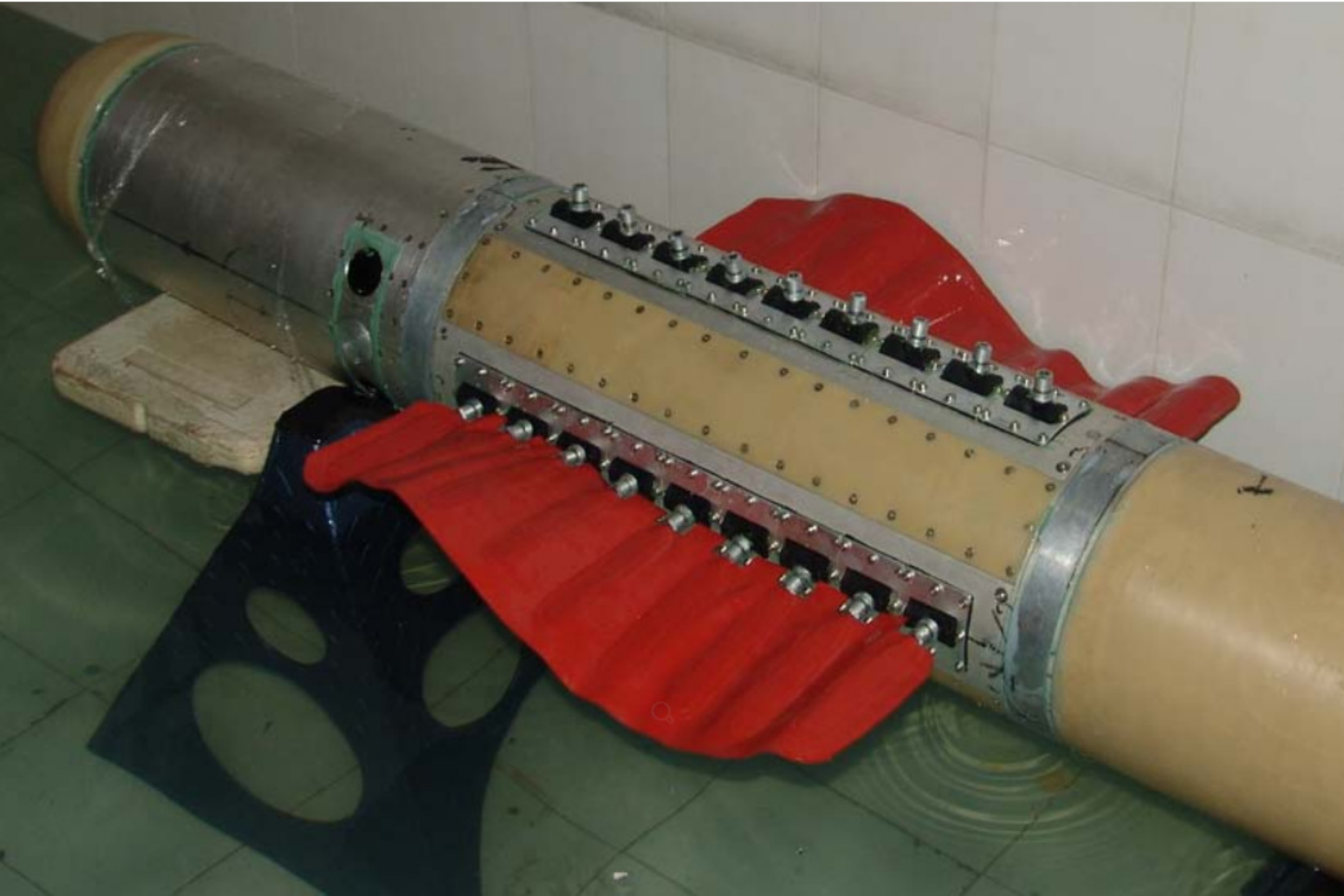} & \includegraphics[scale=0.09]{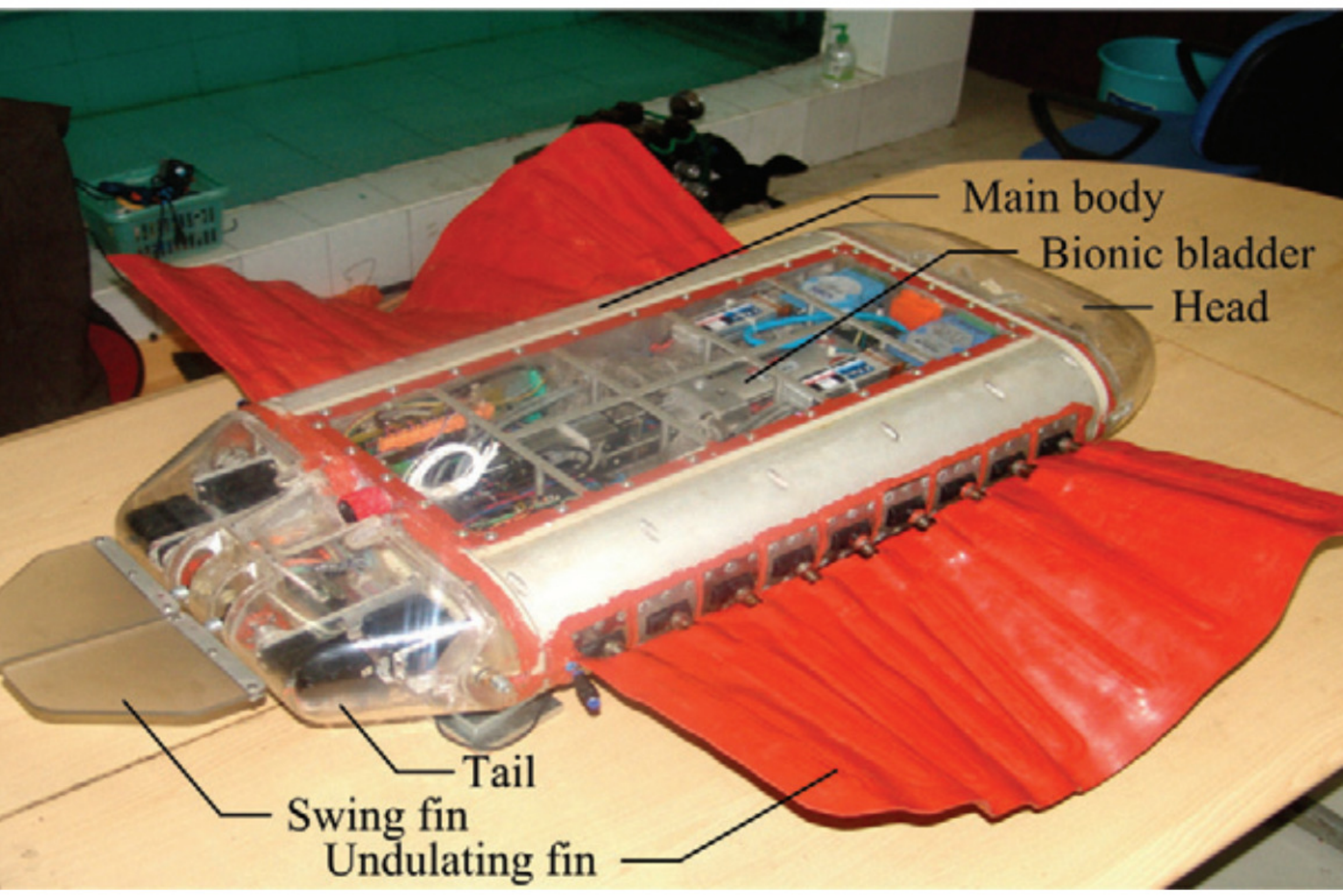} \\
{\footnotesize\sf (a)} & {\footnotesize\sf (b)} \\
\end{tabular}
\caption{Fish-like AUVs used in \citep{lin2009application} (a), and \citep{lin2010supervised} (b).}
\label{fig:fish}
\end{figure}

For fish-like robots, the fin-type propulsive forces and moments depend on integrated influence of various factors such as waveform, wavelength, amplitude and frequency. In \citet{lin2009application}, online Q-learning method was implemented on a bionic underwater robot (Fig.~\ref{fig:fish}a) to select frequencies for its two undulating fins in the autonomous heading control task. The experiment result was barely satisfactory, having relatively big error in yaw angle. In 2010, the same task was carried out on a similar robot with a more flattened body (Fig.~\ref{fig:fish}b), utilizing an improved Q-learning method for continuous state space \citep{lin2010supervised}. The proposed algorithm stored experiences in a replay buffer and removed old experiences according to the resembling degree. A PID controller was adopted for supervising to prevent the occurrence of low learning rate when starting from scratch. The bionic AUV swam smoothly under the modified algorithm. \citet{wang2015optimization} showed that a hierarchical RL structure can enhance the convergence rate of Q-learning in such locomotion problem.

\begin{figure}[t!]
	\centering
	\includegraphics[scale=0.22]{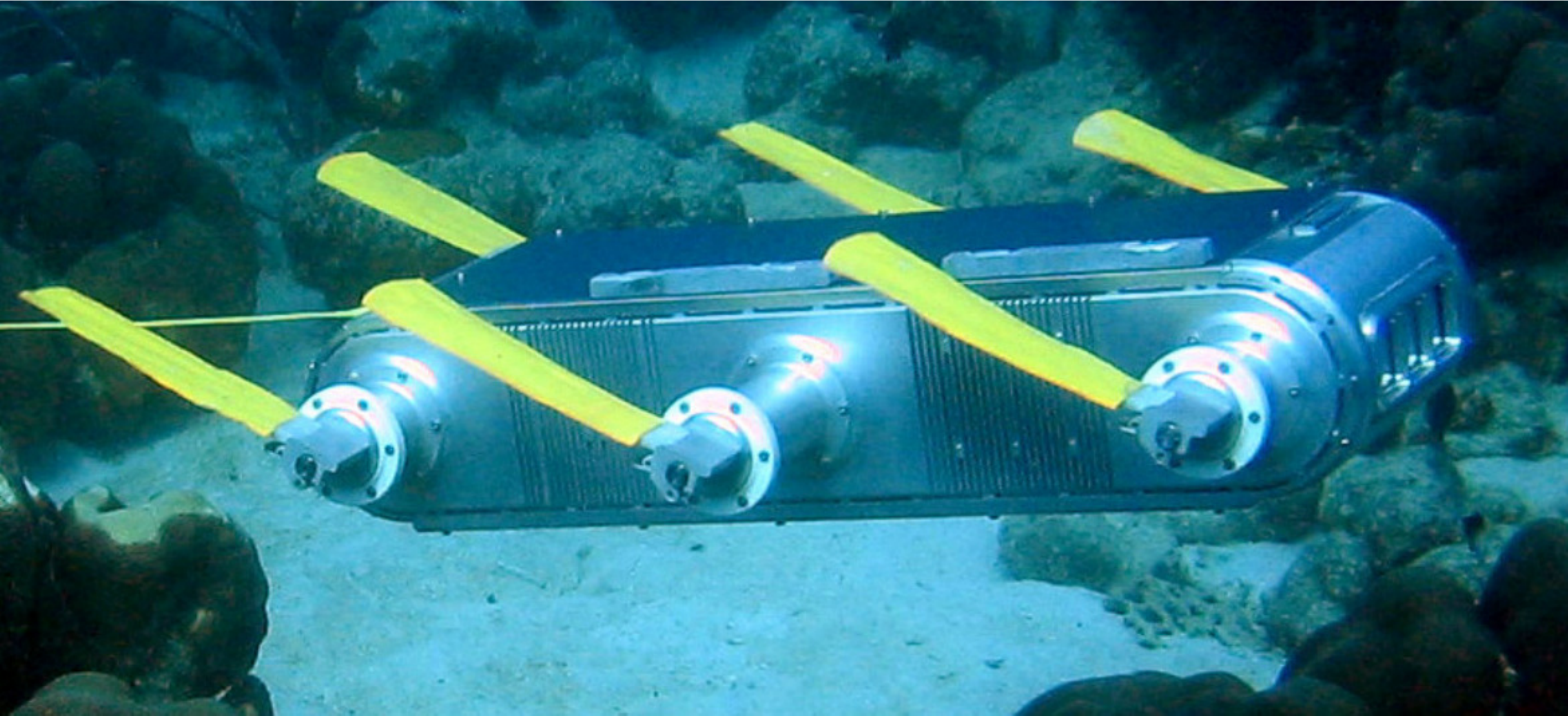}
	\caption{The bionic AUV Aqua \citep{prahacs2004towards}.}
\label{fig:aqua}
\end{figure}

Aqua, as shown in Fig.~\ref{fig:aqua}, is a descendant of hexapod walking vehicle, which has the ability to work underwater \citep{prahacs2004towards}. \citet{meger2015learning} aimed to learn the gait of its six flippers through a policy search method PILCO (Probabilistic Inference for Learning Control), in which a probabilistic dynamic model was learnt before implementing tabular-rasa. Five out of the six different fixed-depth tasks carried out on real robot obtained satisfactory results within seven iterations. Additional experiments about sharing experiences from a simulator showed that an inaccurate model will deteriorate the performance of the proposed method. To address the problem of being computational expensive, \citet{higuera2018synthesizing} proposed an improved deep-PILCO method, which gained competitive data-efficiency while optimizing neural network controllers.

\begin{figure}[tbh]
	\centering
	\includegraphics[scale=0.17]{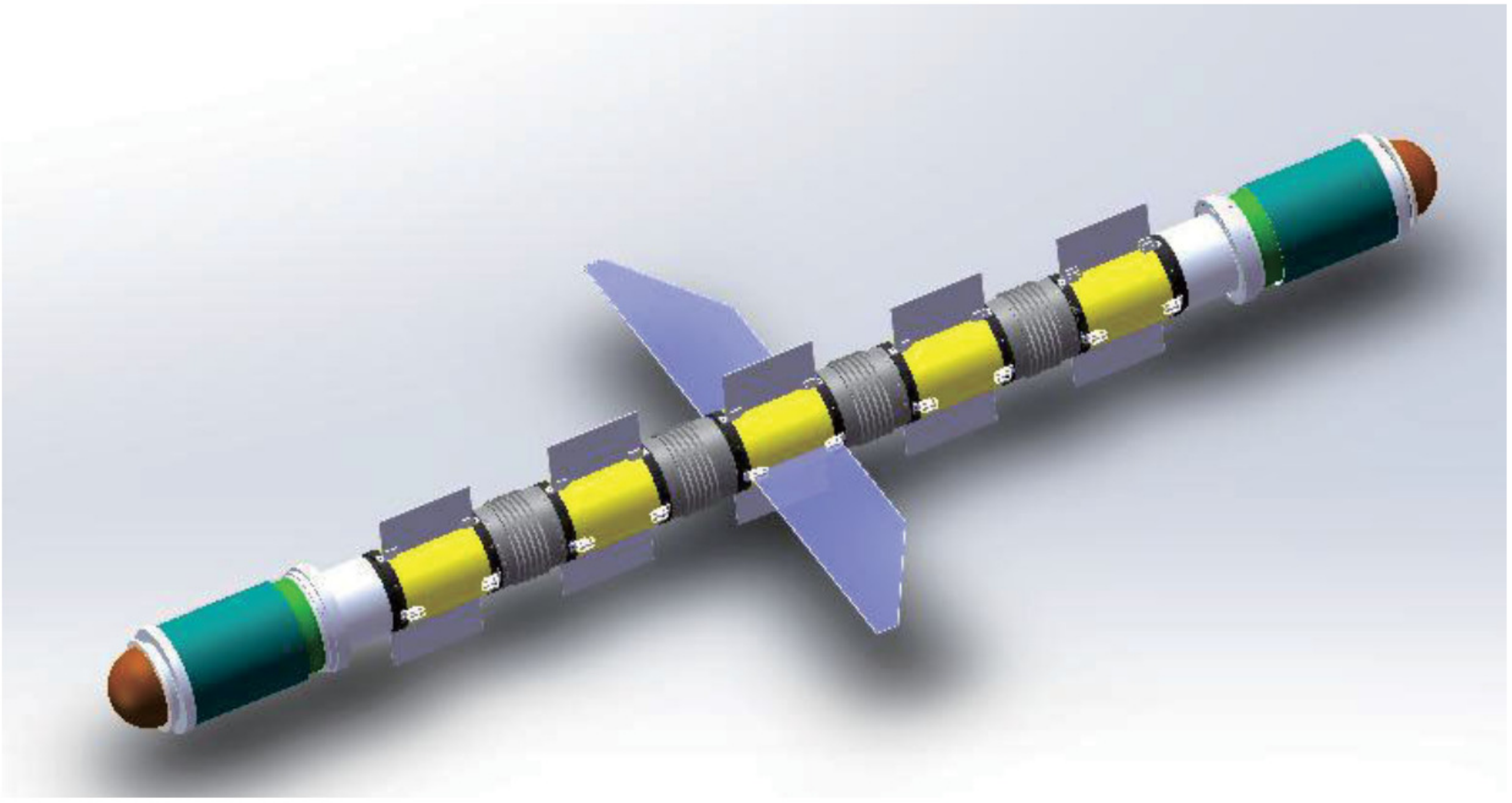}
	\caption{The snake-like AUV used in \citep{zhang2018gliding}.}
\label{fig:snake}
\end{figure}

\citet{zhang2018gliding} concentrated on the control of a snake-like underwater robot that incorporated advantages of the underwater glider through two gliding wings, as shown in Fig.~\ref{fig:snake}. REINFORCE algorithm using preprocessed input was adopted and the  simulation result was encouraging.

\section{Case Study}
In this section, two representative examples of RL-based controllers are introduced in detail. The first task is a regular tracking problem based on a low-level representation of the system state. It provides an alternative approach to formulate an underwater control task. The second task unfolds more possibilities of RL-based controllers by performing end-to-end learning.

\subsection{Seafloor tracking problem}
\subsubsection{Problem formulation}
\begin{figure}[tbh]
	\centering
	\includegraphics[scale=0.15]{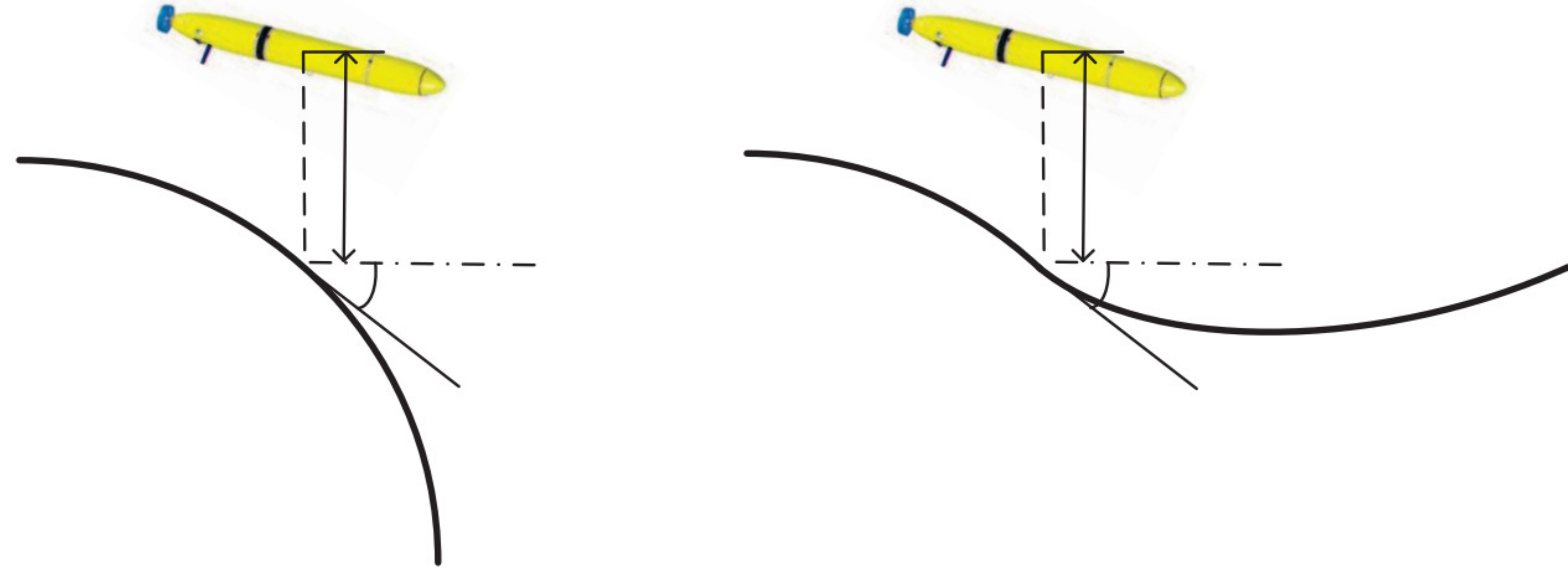}
	\caption{Perceptual aliasing in the depth control problem.}
\label{fig:depthcontrol}
\end{figure}

In a seafloor tracking task, an AUV should keep a certain tracking velocity while holding a constant relative distance $z_r$ with the seafloor. Generally, only motions in vertical plane are considered, in which the surge speed is assumed to be constant. The actions are continuous inputs of the related thrusters and the state of the AUV can be described as $\chi=[z, \theta, w, q]^{T}$, including heave position $z$, heave velocity $w$, pitch orientation $\theta$ and pitch angular velocity $q$. To avoid the confusion due to the periodicity of angle, $[\cos (\theta), \sin (\theta)]^{T}$ is used instead of $\theta$. Moreover, replacing $z$ with a goal oriented variable $\Delta z \doteq z-z_{r}$ can enhance the generality of the learnt policy. Thus, the state can be designed as
\begin{align}
  s=[\Delta z, \cos (\theta), \sin (\theta), w, q]^{T}.
\end{align}
However, as illustrated in Fig.~\ref{fig:depthcontrol}, perceptual aliasing may appear owing to the unknown future trend of the target depth. Whilst this trend is unaccessible in the seafloor tracking problem, it can be predicted by the sequence of recent observations $\left[\Delta z_{t-N+1}, \ldots, \Delta z_{t-1}, \Delta z_{t}\right]$, where $N$ denotes the length of the sequence. In conclusion, the state of the seafloor tracking problem is designed as
\begin{align}
  s=[\Delta z_{t-N+1}, \ldots, \Delta z_{t-1}, \Delta z_{t}, \cos (\theta), \sin (\theta), w, q]^{T}.
\end{align}
The reward is straightforward and given as follows:
\begin{align}
  r=\rho_{1}{\Delta z_{t}}^{2}+\rho_{2} w^{2}+\rho_{3} q^{2}+u^{T}Ru
\end{align}
where the first term aims to minimize the depth error and other terms are for the minimization of the consumed energy. The coefficients can provide tradeoffs among different objectives.

\subsubsection{Methods and strategies}
\begin{figure}[!htb]
\centering
\begin{tabular}{c}
\includegraphics[scale=0.18]{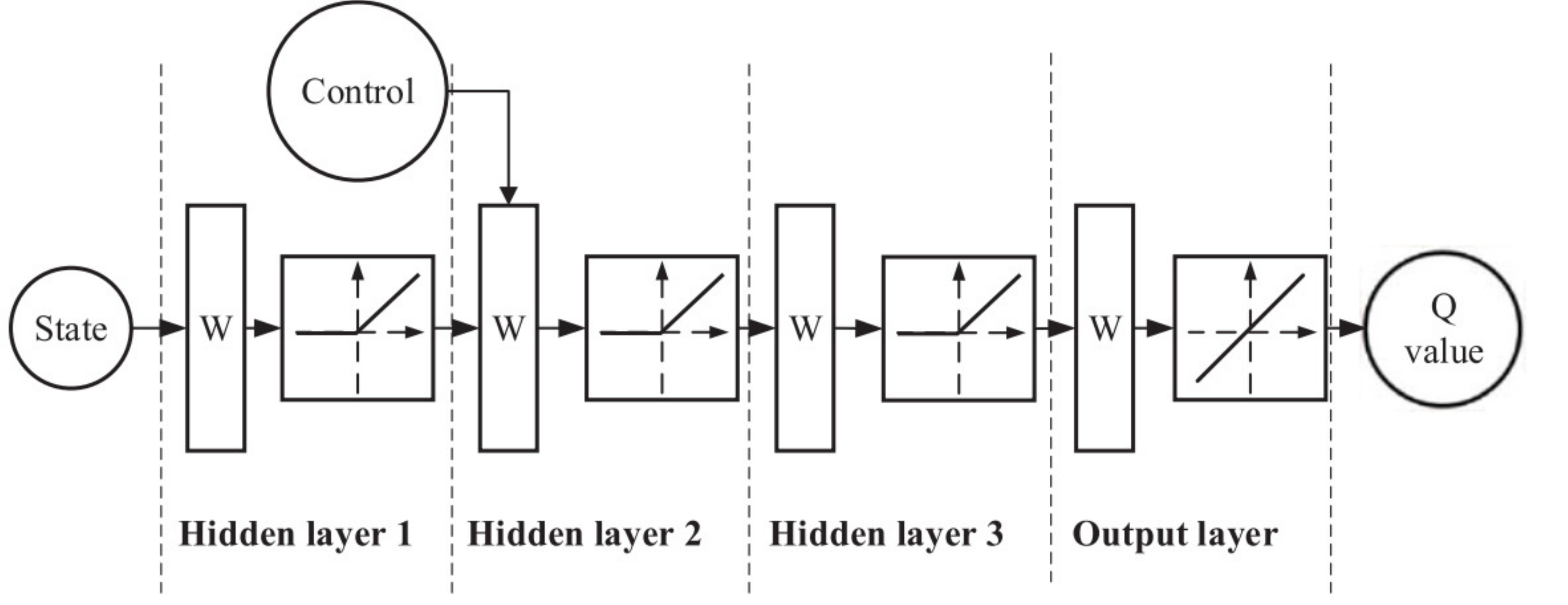}\\
{\footnotesize\sf (a)} \\[3mm]
\includegraphics[scale=0.18]{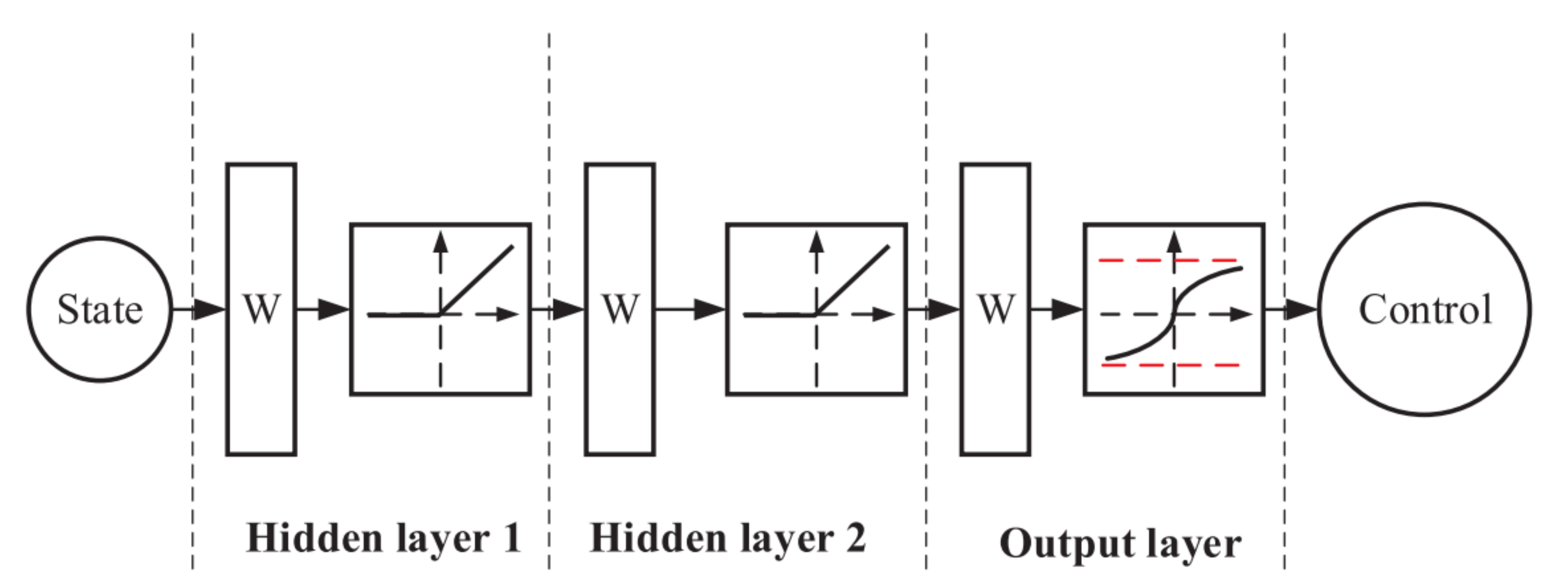}\\
{\footnotesize\sf (b)} \\
\end{tabular}
\caption{Structure of the evaluation network (a), and policy network (b).}
\label{fig:DPGstructure}
\end{figure}

This problem is solved by implementing the DPG algorithm. As is mentioned above, it updates the parameterized policy $\pi_{\theta}$ along the gradient of the performance function $\nabla_{\theta} J\left(\pi_{\theta}\right)$. This gradient is approximated by
\begin{align}
  \nabla_{\theta} J(\theta) \approx \frac{1}{M} \sum_{i=1}^{M} \nabla_{\theta} \pi\left(s_{i} | \theta\right) \nabla_{u_{i}} Q\left(s_{i}, u_{i} | \omega\right)
\end{align}
in which $\left(s_{k}, u_{k}, s_{k+1}\right)$ is a transition pair along a trajectory at time $k$ and $Q(s, u | \omega)$ is a parameterized approximation for the value function. As illustrated in Fig.~\ref{fig:DPGstructure}, both the policy and value function are approximated by neural networks(NNs), with three layers and four layers respectively. The activation function ReLu is used for better convergence rate. To improve sample efficiency, prioritized experience replay \citep{schaul2015prioritized}, which reuses previous experiences according to their priority, is adopted. The priority of an experience is proportional to its TD error
\begin{align}
  \mathrm{PRI}_{k}=| r_{k}+\gamma Q\left(s_{k+1}, \pi\left(s_{k+1} | \theta\right)| \omega\right)-Q\left(s_{k}, u_{k} | \omega\right)|.
\end{align}
The intuition behind this definition is that a RL agent can learn more from a transition with higher magnitude of TD error. During the training, samples with higher priority are more likely to be chosen.

\subsubsection{Results}
\begin{figure}[tbh]
	\centering
	\includegraphics[scale=0.18]{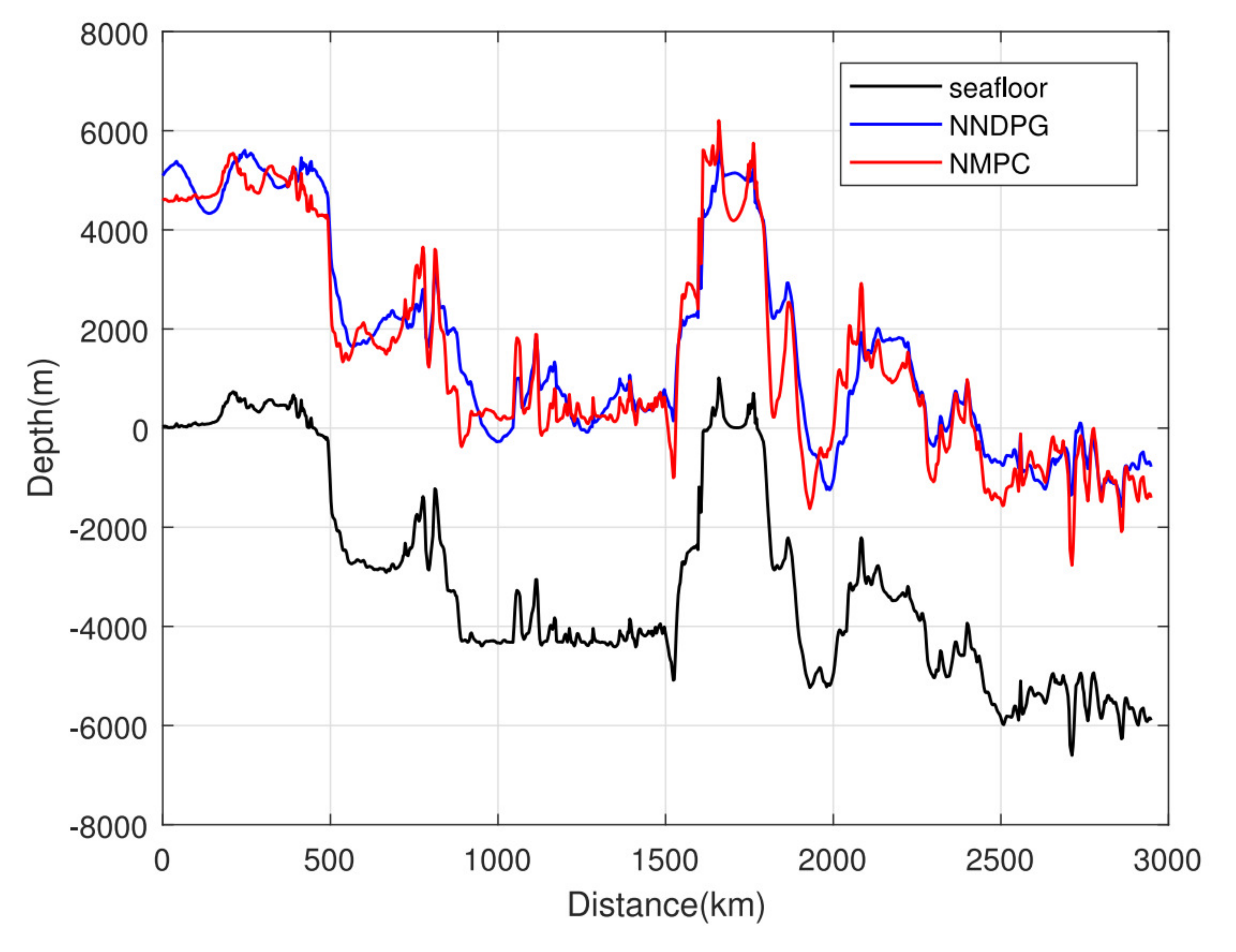}
	\caption{Tracking trajectory of NNDPG, NMPC, and the realistic seafloor.}
\label{fig:depthresult}
\end{figure}

Simulation tests are carried out on a path generated by a data set sampled from the real seafloor of the South China Sea at $(23\degree06'N, 120\degree07'E)$, which is provided by the Shenyang Institute of Automation, Chinese Academy of Science. The number of the preserved $\Delta z$ is three, which is decided by preliminary experiments. Fig.~\ref{fig:depthresult} shows that the proposed controller performs well in the test and is comparable with nonlinear model predictive control (NMPC) without having to know the dynamics of the AUV.

\subsection{End-to-end control problem}
\subsubsection{Problem formulation}
\begin{figure}[!htb]
\centering
\begin{tabular}{cc}
\includegraphics[scale=0.13]{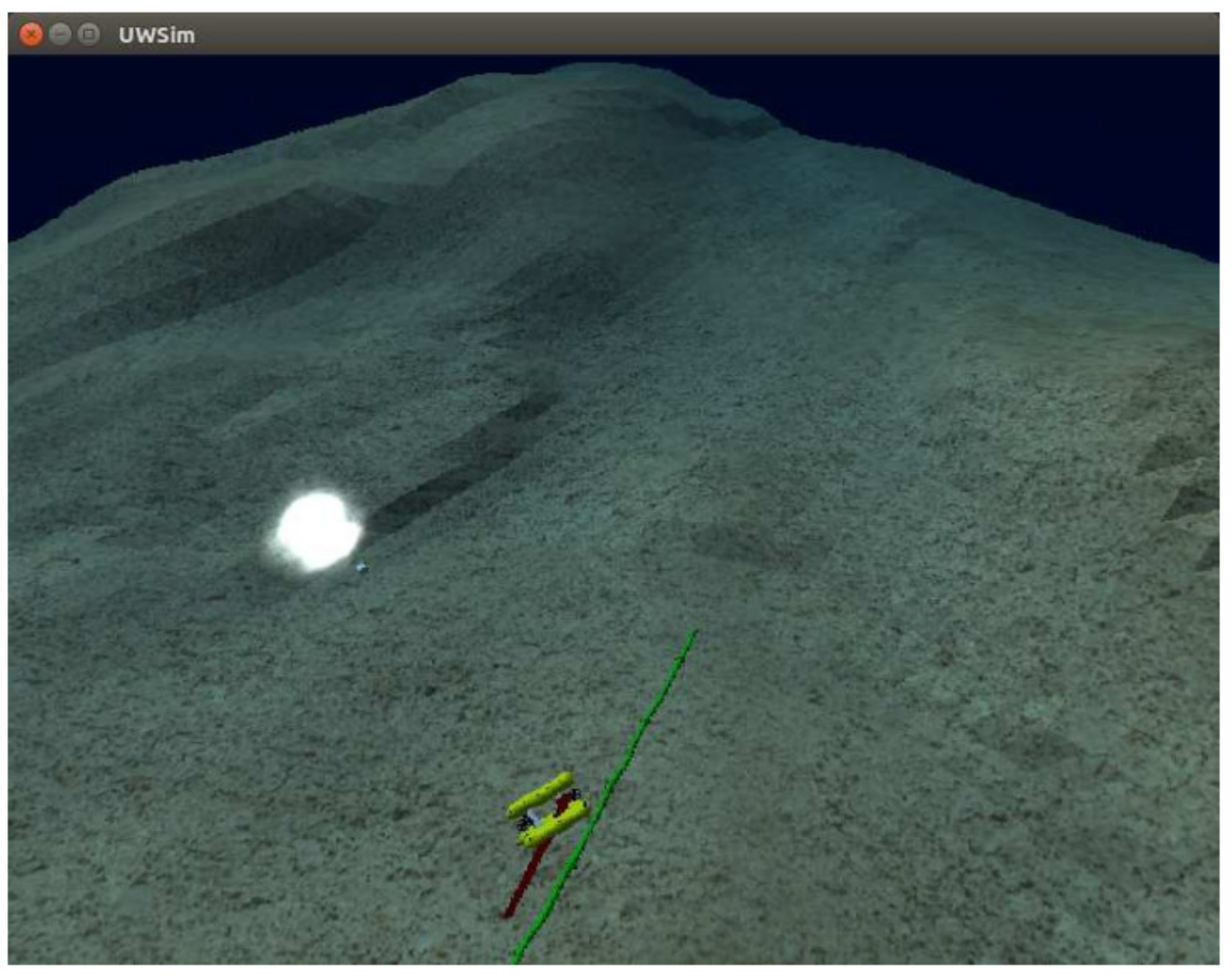} & \includegraphics[scale=0.13]{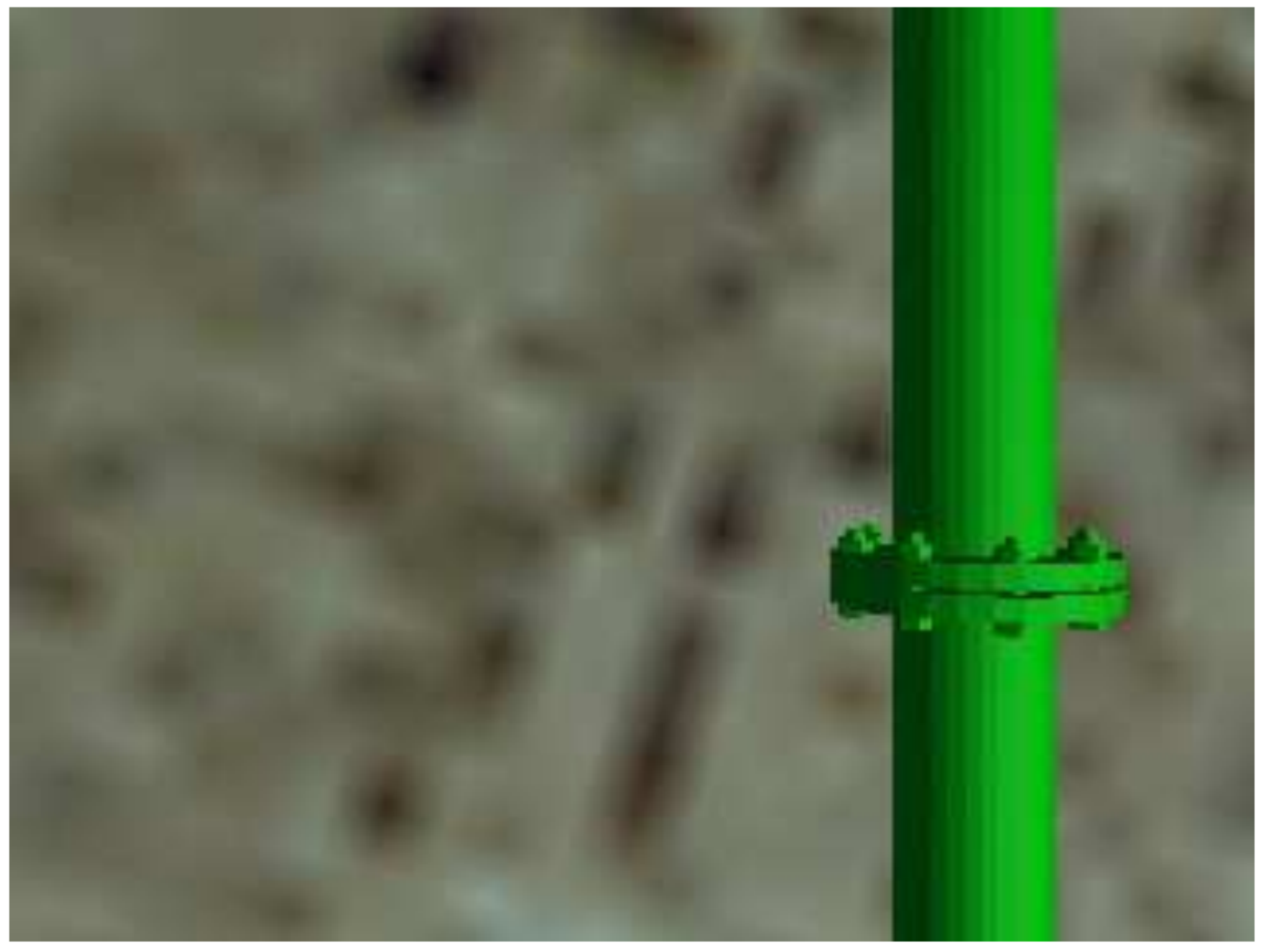} \\
{\footnotesize\sf (a)} & {\footnotesize\sf (b)} \\
\end{tabular}
\caption{Simulation scene for pipe following (a), and view of the camera (b).}
\label{fig:scene}
\end{figure}

Deep learning is capable of learning from unprocessed, high-dimensional and sensory input. In other words, it has the ability to construct end to end solution, which is preferable in most situation. On the other hand, algorithms with this kind of inputs are usually hard to converge, especially when dealing with low level control problems involving complex dynamics. Here we present an example which proposed an end-to-end control policy for the pipe following task using sensor signals and motion variables as inputs. The AUV has to keep the pipeline in its camera view and head along the pipeline without knowing its own position as well as that of the pipeline (Fig.~\ref{fig:scene}). The sensor input is an 84x84x3 image and the motion variables contain the orientation vector and velocity vector.

\begin{figure}[tbh]
	\centering
	\includegraphics[scale=0.25]{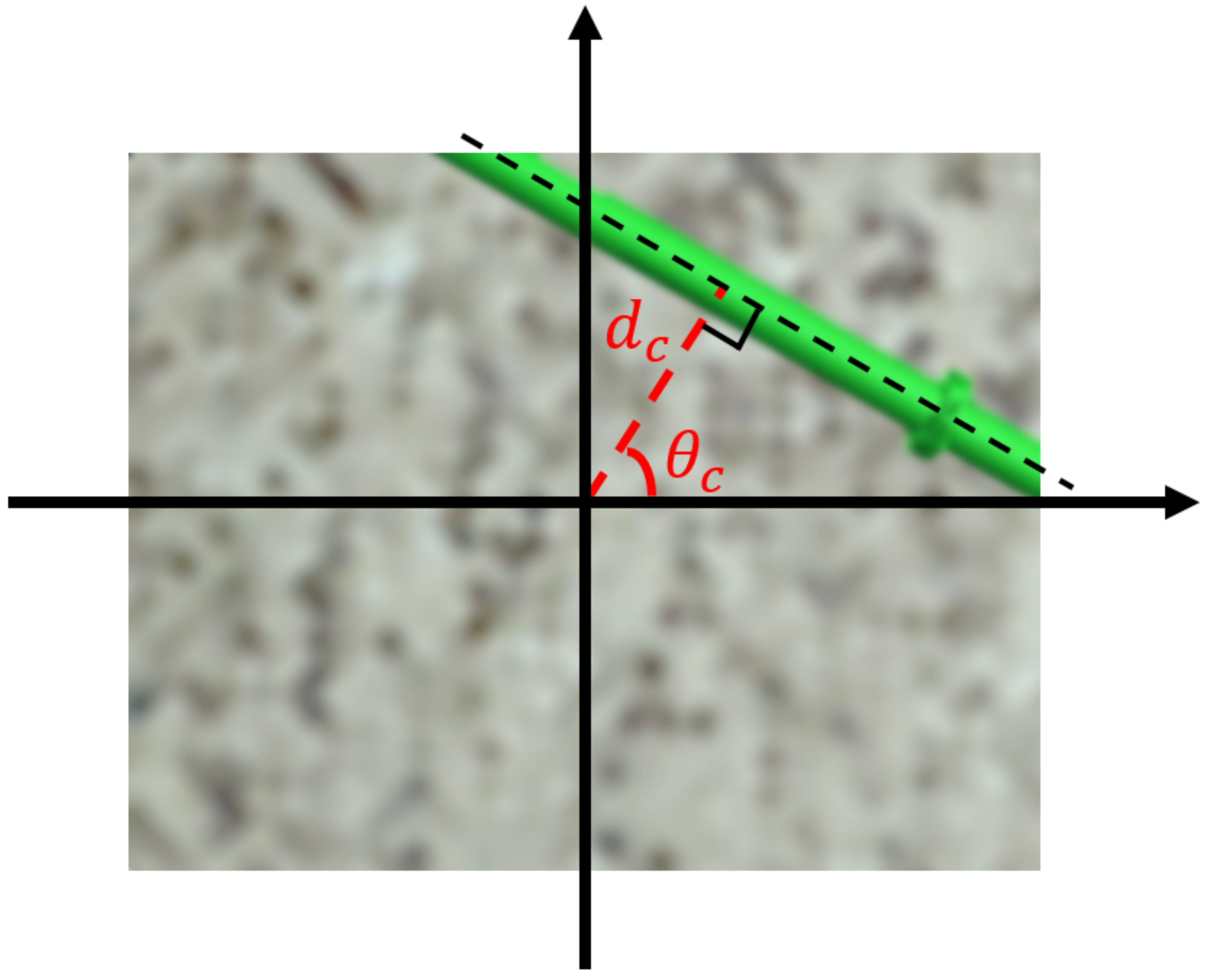}
	\caption{Illustration of $d_{c}$ and $\theta_{c}$.}
\label{fig:pipe}
\end{figure}

Although the controller is trained directly through the raw image input, image processing is leveraged to aid the reward extraction. After a series of procedures, the center line of the pipeline in the camera view can be detected. Its distance from the center of the view $d_{c}$ and angle between the distance line and x axis $\theta_{c}$ are then calculated (Fig.~\ref{fig:pipe}). The reward is designed as
\begin{align}
  r=u \cdot\left(\left|\cos \theta_{c}\right|-d_{c}d_{\max }^{-1}\right)
\end{align}
where $u$ denotes the surge velocity and $d_{\max }$ equals half of the diagonal length of the view.
The actions are the input of the two related thrusters.

\subsubsection{Methods and strategies}
\begin{figure}[tbh]
	\centering
	\includegraphics[scale=0.14]{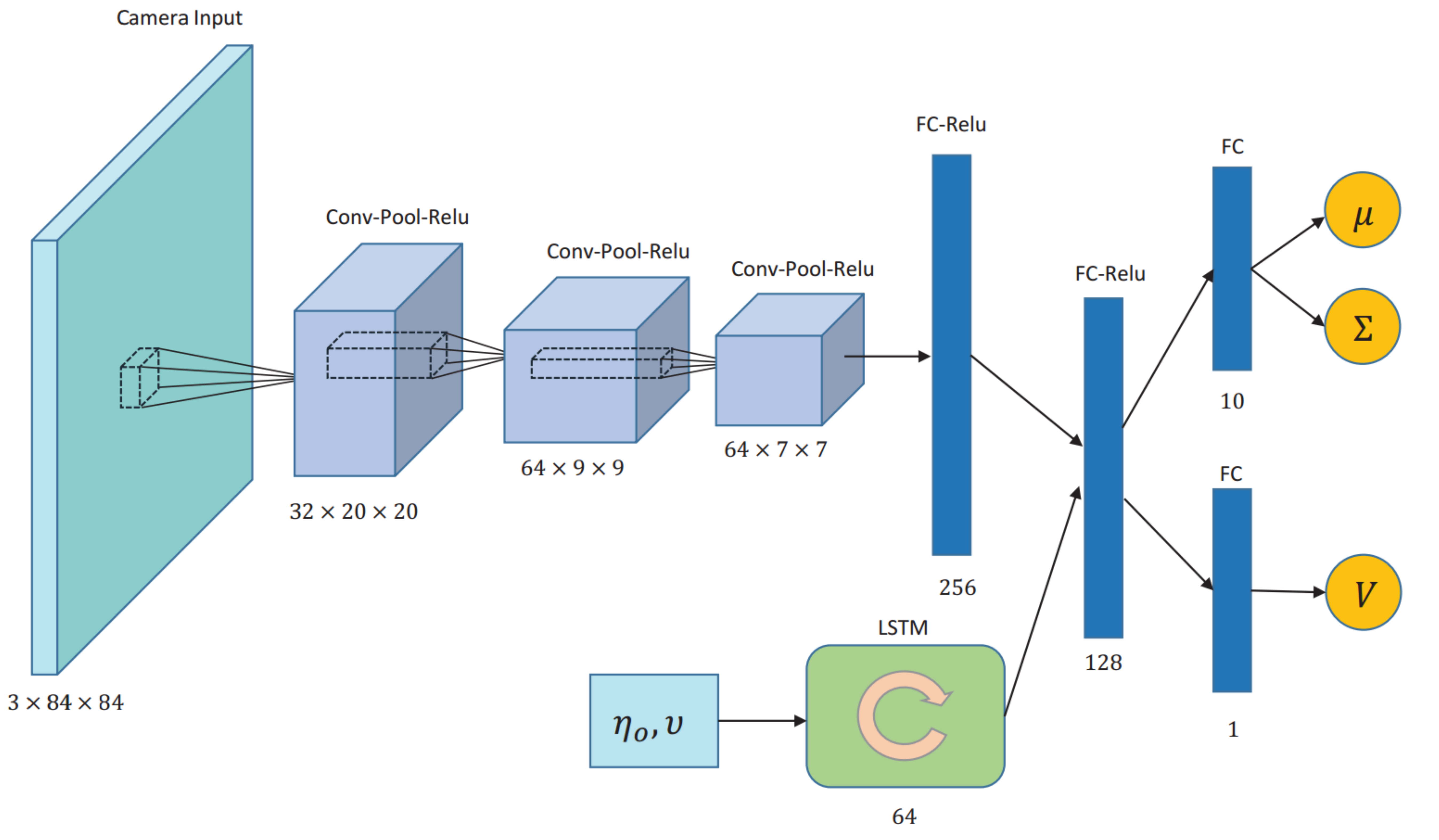}
	\caption{Structure of the network.}
\label{fig:ppo}
\end{figure}

The sensor input and motion variables are handled with two encoder networks respectively. The former is a four-layers CNN network and the latter is a LSTM(long short term memory) network. As illustrated in Fig.~\ref{fig:ppo}, both of their outputs are fed to a fully connected layer followed by a value network and a policy network. Proximal policy optimization, a kind of natural policy gradient method, is implemented for training the network. Besides using the performance function that is defined directly as the cumulative reward, the objective of PPO is given as
\begin{equation}
	\begin{aligned} L^{PPO}=& L^{CLIP}-\lambda_{1} \hat{\mathbb{E}}_{t}\left[\left(V_{\theta}\left(s_{t}\right)-V_{t}^{\operatorname{targ}}\right)^{2}\right] \\ &+\lambda_{2} \hat{\mathbb{E}}_{t}\left[H\left(\pi_{\theta}\left(\cdot | s_{t}\right)\right)\right] \end{aligned}
\end{equation}
where $\hat{\mathbb{E}}_{t}$ means to average over a batch of samples. The second term minimizes the error of the value function for better estimation of the advantage function $\hat{A}_{t}$  and the last term encourages exploration. The advantage function measures the relative advantage of an action and is mathematically defined as $A^{\pi}(s, a)=Q^{\pi}(s, a)-V^{\pi}(s)$. $V_{t}^{\operatorname{targ}}$ can be the cumulative return and $H$ computes the entropy of a distribution.
$L^{CLIP}$ is a clipped surrogate objective
\begin{align}
    L^{CLIP}= \hat{\mathbb{E}}_{t}&\left[\min \left(w_{t}(\theta) \hat{A}_{t}, \notag\right.\right.\\
    \phantom{=\;\;}&\left.\left. \operatorname{clip}\left(w_{t}(\theta), 1-\epsilon, 1+\epsilon\right) \hat{A}_{t}\right)\right]
\end{align}
where the clip function $clip(x,a,b)$ restricts $x$ to the bound $[a,b]$ and $w_{t}(\theta)$ is a weight that measures the difference between policies by calculating
\begin{align}
  w_{t}(\theta)=\frac{\pi_{\theta}\left(a_{t} | s_{t}\right)}{\pi_{\theta_{\text {old}}}\left(a_{t} | s_{t}\right)}.
\end{align}

\subsubsection{Results}
\begin{figure}[tbh]
	\centering
	\includegraphics[scale=0.18]{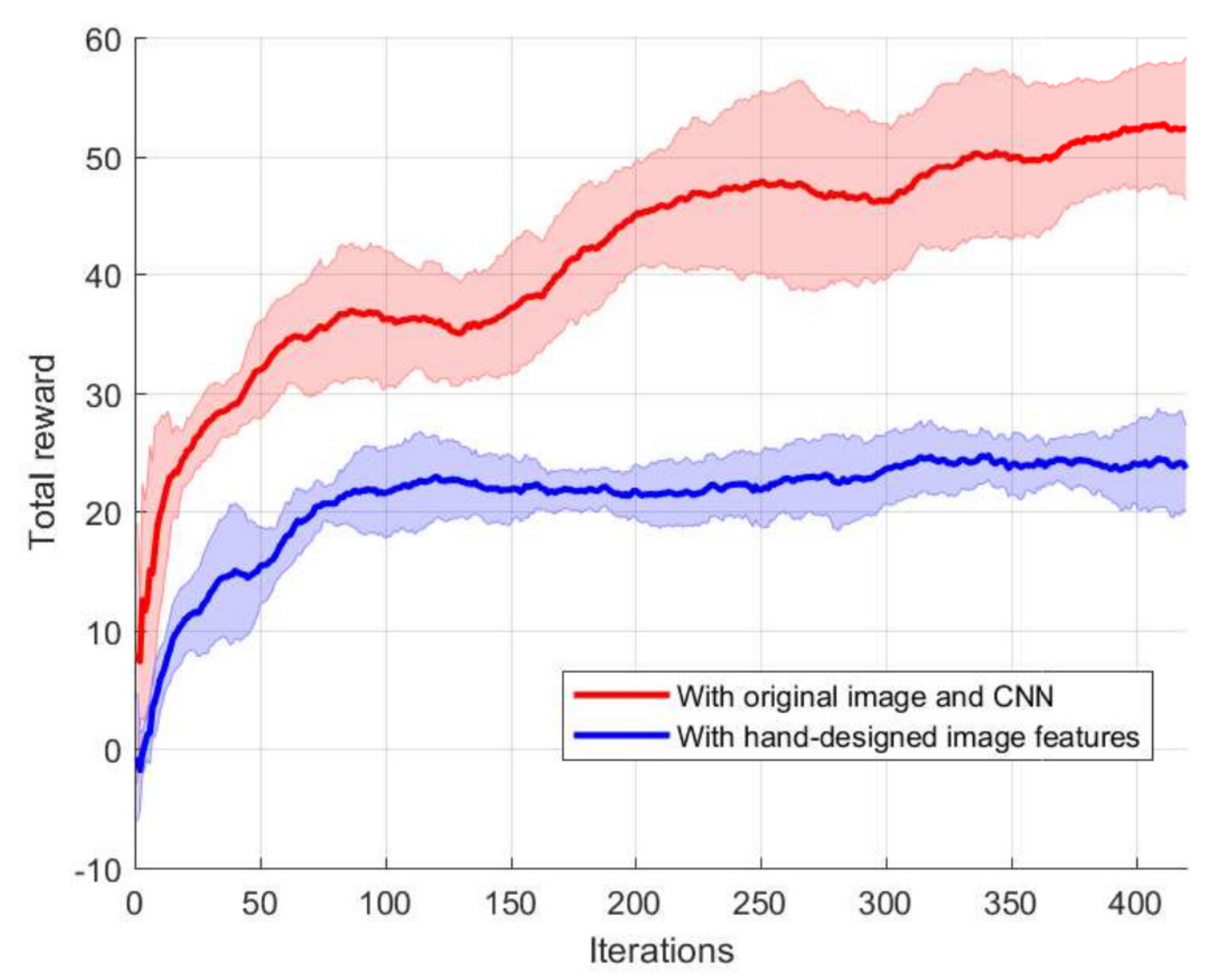}
	\caption{Comparison between the use of hand-designed image features and the CNN encoder.}
\label{fig:handdesign}
\end{figure}

\begin{figure}[tbh]
	\centering
	\includegraphics[scale=0.15]{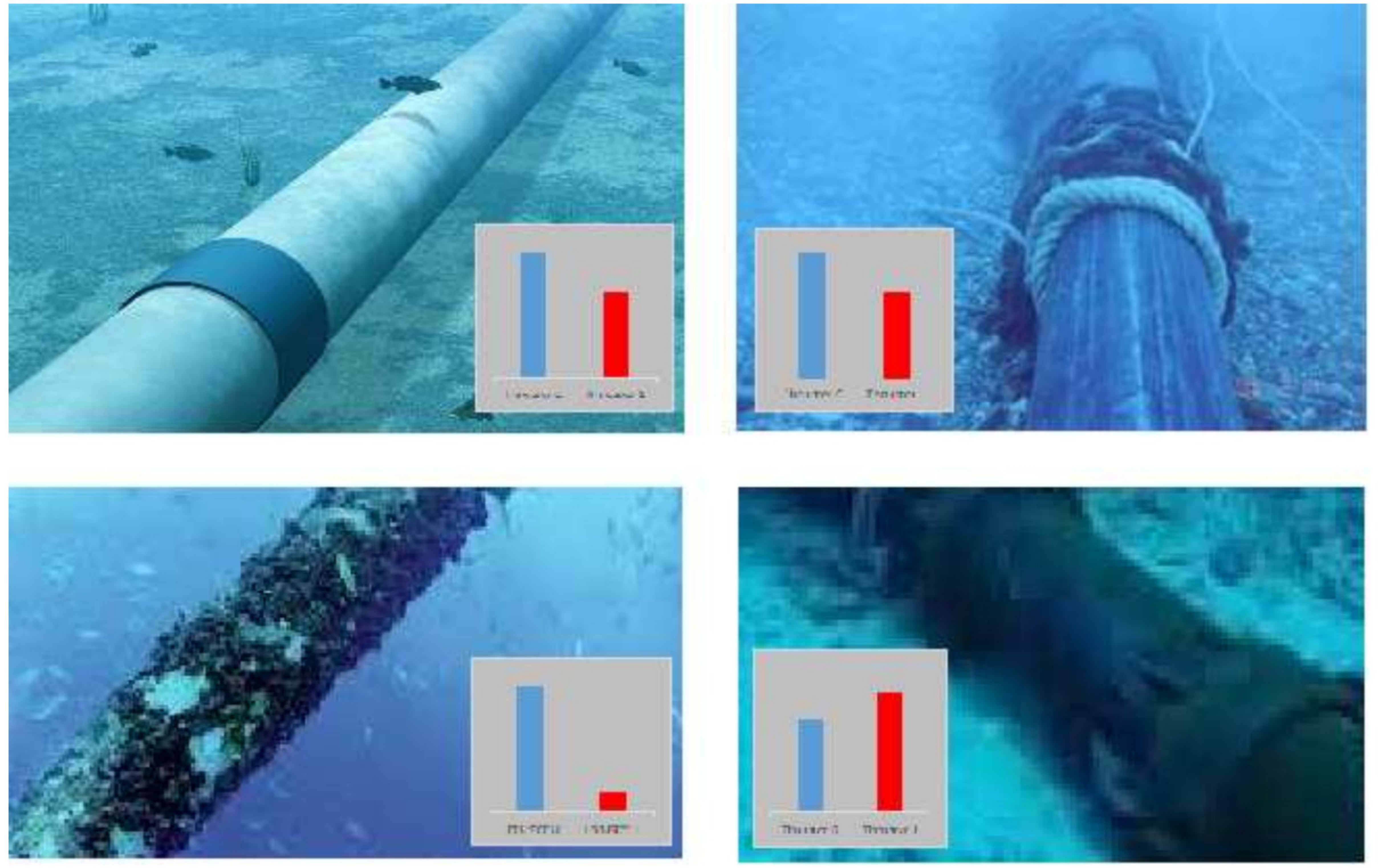}
	\caption{The inference of the learnt policy on several realistic underwater pipeline images. When the output of the left thruster (blue bar) is larger than that of the right thruster (red bar), the AUV tends to turn right, vice versa.}
\label{fig:realpipe}
\end{figure}

The designed controller behaves well on the pipeline tracking task in the simulation scene. The AUV follows the straight pipeline successfully without requiring its localization information and dynamic model. Two extra experiments highlight the advantage of using end-to-end control structure and the generality of the learnt policy respectively. The first experiment replaces the CNN network with the extracted features $\theta_{c}$ and $d_{c}$. Other settings remain unchanged. The results in Fig.~\ref{fig:handdesign} show that the network with CNN performs much better, which indicates that the use of the raw sensory input helps preserve more useful information. The other experiment checks the predicted actions when the sensory input changes from views of the simulated scene to images of realistic underwater pipelines (Fig.~\ref{fig:realpipe}). Actions generated from 21 out of 30 images move the AUV towards the correct direction. Though the magnitude of these actions are barely satisfactory, the results imply the potential of the algorithm to be applied in real world training.

\section{Conclusion}
This paper provides a selective overview of controlling AUVs with RL. Methods that have been proposed in the literature are presented according to the motion control task they are designed for. Whilst there are still many challenges for merging these two areas, steady progress is being made to gain an optimal and practical solution via RL. Furthermore, we list two detailed cases to help to reveal the feasibility and potential of RL in the underwater control domain. We believe that RL-based controllers can pave the way to more intelligent AUVs.


\end{document}